\PassOptionsToPackage{table}{xcolor}
\documentclass[10pt,twocolumn,letterpaper]{article}

\usepackage[pagenumbers]{cvpr} % To force page numbers, e.g. for an arXiv version
\usepackage{multirow}

\usepackage[dvipsnames]{xcolor}

\definecolor{cvprblue}{rgb}{0.21,0.49,0.74}
\usepackage[pagebackref,breaklinks,colorlinks,citecolor=cvprblue]{hyperref}

\def\paperID{*****} % *** Enter the Paper ID here
\def\confName{CVPR}
\def\confYear{2024}

\title{VolETA: One- and Few-shot Food Volume Estimation}

\author{Ahmad AlMughrabi\\
Universitat de Barcelona, Spain\\
{\tt\small sci.mughrabi@gmail.com}
\and 
Umair Haroon \\
Universitat de Barcelona, Spain\\
{\tt\small umairharoon3797@gmail.com}
\and
Ricardo Marques\\
Universitat de Barcelona, Spain\\
Computer Vision Center, Barcelona\\
{\tt\small ricardo.marques@ub.edu}
\and 
Petia Radeva\\
Universitat de Barcelona, Spain\\
Institut de Neurosciències, Barcelona\\
{\tt\small petia.ivanova@ub.edu}
}

\begin{document}
\maketitle
\begin{abstract}
Accurate food volume estimation is essential for dietary assessment, nutritional tracking, and portion control applications. We present VolETA, a sophisticated methodology for estimating food volume using 3D generative techniques. Our approach creates a scaled 3D mesh of food objects using one- or few-RGBD images. We start by selecting keyframes based on the RGB images and then segmenting the reference object in the RGB images using XMem++. Simultaneously, camera positions are estimated and refined using the PixSfM technique. The segmented food images, reference objects, and camera poses are combined to form a data model suitable for NeuS2. Independent mesh reconstructions for reference and food objects are carried out, with scaling factors determined using MeshLab based on the reference object. Moreover, depth information is used to fine-tune the scaling factors by estimating the potential volume range. The fine-tuned scaling factors are then applied to the cleaned food meshes for accurate volume measurements. Similarly, we enter a segmented RGB image to the One-2-3-45 model for one-shot food volume estimation, resulting in a mesh. We then leverage the obtained scaling factors to the cleaned food mesh for accurate volume measurements. Our experiments show that our method effectively addresses occlusions, varying lighting conditions, and complex food geometries, achieving robust and accurate volume estimations with 10.97\% MAPE using the MTF dataset. This innovative approach enhances the precision of volume assessments and significantly contributes to computational nutrition and dietary monitoring advancements. The source code is accessible at\footnote{https://github.com/GCVCG/VolETA-MetaFood}.
\end{abstract}    
\section{Introduction}
\label{sec:intro}
Photographs, typically 2D digital images, are extensively employed to gather data on dietary intake \cite{dehais2016two, frobisher2003estimation}. In studies conducted in naturalistic settings, participants frequently capture images before, during, or after meals to document food consumption. These photographs are subsequently analyzed to classify eating behaviors, including identifying food types and assessing nutritional intake \cite{amoutzopoulos2020portion}.

Accurate nutrient analysis of food, however, often necessitates the determination of food weight, which is derived from both the density and volume of the items \cite{liu2020food}. Food density can be estimated by identifying and applying standard food density tables, though these tables are only available for a limited range of foods \cite{lo2020image}. Estimating volume from 2D images is inherently challenging without 3D depth information, which standard mobile phone cameras do not typically provide \cite{dehais2016two, okamoto2016automatic}. Consequently, ongoing research is focused on developing methodologies to derive precise 3D volume estimates from 2D images \cite{ho2021integration}.

There are two principal approaches to food volume estimation: those utilizing reference objects in images \cite{liu2020food,okamoto2016automatic, smith2022food} and those that do not \cite{puri2009recognition, steinbrener2023learning}. Reference objects, which can be explicitly placed before image capture — such as a card \cite{okamoto2016automatic, smith2022food}, local currency or a standard-sized object like a Rubik’s cube \cite{liu2020food} — or extracted from identified image elements like plates or serving vessels with known dimensions \cite{beltran2019reliability}, provide 3D cues necessary for depth estimation \cite{dehais2016two}. Alternatively, methods without reference objects attempt to determine image depth by identifying features such as parallel elements that can establish a vanishing point or by leveraging properties of the camera placement and environmental context \cite{puri2009recognition,yang2019image,steinbrener2023learning}, effectively using the camera as its reference object.

Approaches involving reference objects can be computationally intensive, often requiring the deployment of neural networks  \cite{liu2020food,okamoto2016automatic,steinbrener2023learning,thames2021nutrition5k} or multiple images of the food item \cite{dehais2016two,thames2021nutrition5k}. Additionally, reference and non-reference object methods face significant challenges due to the casual nature of 2D image capture. Cameras are typically not mounted on tripods, resulting in variable orientations and positions that are not accurately recorded. A recent hybrid approach \cite{yang2019image} proposes grounding mobile phones on a flat surface and incorporating camera orientation data as machine-readable metadata within the image capture. However, this technique is constrained by the need for a stable surface, which may not always be available, particularly in rural or developing regions. Furthermore, strict placement requirements can be inconvenient for users, who may forget to adhere to these protocols consistently, compromising image capture accuracy. Our approach provides a semi-automated method for estimating food volume to address the following challenges:
\begin{itemize}
    \item Unbounded food scenes, where the camera can freely move around the food object. For instance, the capturing might have different capturing speeds and topologies.   
    \item Sparse input views, where food scenes may consist of just one or a few RGBD images. 
\end{itemize}
The remainder of this document is structured as follows: we will present our related work in Sec. \ref{sec:related_work}; in Sec.~\ref{sec:methodology}, we present our detailed methodology; in Sec. ~\ref{sec:results}, we present our qualitative and quantitative results; and finally, in Sec.~\ref{sec:conclusion}, we present our conclusion and future work.
\section{Related work}
\label{sec:related_work}

While recent advancements in computer vision and deep learning have assisted the capabilities of automated food volume estimation systems, early methods were heavily reliant on traditional image processing techniques, such as contour detection \cite{gao2019musefood} and shape analysis \cite{gao2018food}. These methods often necessitated user input and were hampered by their dependence on simplistic models that could not generalize well to diverse food types.

Furthermore, some approaches utilize reference objects and rely on known dimensions of objects within the image to infer the volume of food items. For instance, combining  Faster  R-CNN,  Grabcut,  Median filtering,  and  CNN algorithm estimating food volume based on reference standard-sized objects like Rubik's cubes to provide the necessary scale for depth estimation \cite{liu2020food}. Similarly, Okamoto et al. \cite{okamoto2016automatic} and Smith et al. \cite{smith2022food} proposed an image-based volume estimation system using pre-registered reference objects, e.g., cards or local currency beside the food before capturing the image. Also, Beltran et al. \cite{beltran2019reliability} explored common elements like plates or serving vessels with predefined dimensions as implicit reference objects. These methods have demonstrated effectiveness in providing 3D cues, but they often require manual placement of reference objects, which can be cumbersome and prone to user error.

%One prominent line of research focused on 3D reconstruction techniques, utilizing multiple images or depth sensors to model food items. Approaches such as Structure from Motion (SfM) and RGB-D sensing provided more accurate volume estimations by capturing the three-dimensional structure of food items~\cite{jones2018three}. With the advent of deep learning, convolutional neural networks (CNNs) have been employed to estimate food volume from images directly. These methods leverage large datasets and powerful neural architectures to learn features robust to variations in food types, lighting conditions, and presentation styles~\cite{chen2020deep}. Recent advancements have introduced generative adversarial networks (GANs) and multi-view learning approaches, which synthesize realistic 3D models of food items from single or multiple images. Additionally, cutting-edge 3D sensing technologies, such as LiDAR and structured light, have further improved the accuracy and reliability of food volume estimation systems~\cite{wang2019generative,liu2021cutting}. While 3D reconstruction and deep learning methods have propelled the field forward, they often demand extensive computational resources and large annotated datasets. Furthermore, these methods may grapple with occlusion and complex food presentations~\cite{zhang2020challenges}. 

Non-reference object methods attempt to estimate depth by leveraging 3D reconstruction techniques, utilizing multiple images and depth sensors to model food items, or using the camera position as a reference. Puri et al. \cite{puri2009recognition} used camera pose estimation alongside dense stereo matching and 3D reconstruction, and   Yang et al. \cite{yang2019image} proposed a hybrid approach where mobile phones are placed on flat surfaces, and orientation data is included as metadata, which aids in depth estimation. Another innovative method involves learning metric volume estimation from short video sequences consisting of monocular RGB video frames and associated inertial data, i.e., IMU using LSTM, which allows for more dynamic and flexible depth estimation from casual video captures \cite{steinbrener2023learning}. While these methods are innovative, they still face potential data dependency, limited applicability to certain food items, computational complexity, challenges related to accuracy and precision, insufficient validation and benchmarking, potential user interaction issues, and uncertainty regarding real-world deployment feasibility.

Both reference and non-reference object methods encounter substantial computational challenges. Methods involving reference objects often require sophisticated neural networks and multiple image captures to estimate volume accurately. Thames et al. \cite{thames2021nutrition5k} highlighted the computational intensity of neural networks deployed for these tasks, emphasizing the need for powerful hardware such as Intel RealSense/Microsoft Kinect depth sensors and substantial processing time. In casual settings, where cameras are handheld and have varied orientations and positions, these methods struggle to maintain accuracy due to the lack of consistent reference points and stable conditions.

On the other hand, hybrid approaches attempt to balance the advantages and limitations of reference and non-reference methods. Yang et al. \cite{yang2019image} suggested grounding mobile phones on a flat surface to capture orientation data, improving depth estimation accuracy. However, this approach requires a stable surface, which might not always be feasible, especially in less controlled environments like rural or developing regions. Additionally, these techniques depend heavily on user adherence to specific protocols, which can be inconsistent.

% % NeuS2
% Recent advancements in neural surface representation, such as NeuS \cite{wang2021neus}, have achieved high-quality static scene reconstructions but are hindered by long training times (approximately 8 hours), making them unsuitable for dynamic scenes. To overcome this limitation, NeuS2 was developed, offering two orders of magnitude faster training without compromising quality. NeuS2 \cite{wang2023neus2} utilizes multi-resolution hash encodings and lightweight second-order derivative calculations optimized for CUDA parallelism, significantly boosting training speed. It also incorporates a progressive learning strategy for stabilization and efficiency. NeuS2 features an incremental training strategy and a global transformation prediction component for dynamic scenes, effectively handling long sequences with large movements. Experimental results demonstrate that NeuS2 excels in both accuracy and speed for static and dynamic scene reconstruction compared to current state-of-the-art methods.

Our proposed pipeline addresses these challenges by offering a semi-automated food volume estimation approach suitable for unbounded scenes with free camera movement and sparse input views. This method utilizes both one and few-shot RGBD images and seeks to balance accuracy and usability, making it more adaptable to various real-world scenarios. Regarding estimating a few shots of food volume estimation, a few images involve RGBD (color and depth) images and masks that outline the food objects. The main {\bf contributions} of this article are:
\begin{enumerate}
    \item We introduce a framework that effectively handles challenges, including occlusions, varying lighting conditions, and complex food geometries. Our approach achieves robust and accurate volume estimations with a 10.97\% MAPE using the MTF dataset.
    \item  Our framework introduces a new keyframe selection layer to address the ambiguity issue that arises in unbounded scenes. This layer incorporates two optimized filters, Defocus Blur, and near-image Similarity, eliminating noisy, blurry, and redundant input images.
    \item We adopt an advanced and innovative camera pose estimation technique called PixSfM \cite{lindenberger2021pixel} instead of the usually used Colmap tool for Camera Pose Estimation \cite{schonberger2016structure}. PixSfM enables us to obtain more intricate and accurate camera poses and features, which helps any NeRF-like architectures generate highly detailed and photo-realistic renderings of unbounded scenes. We show that our framework based on PixSfM can extract dynamic and high-level features and differentiate between extracted camera positions corresponding to common features, significantly reducing the ambiguity in camera pose estimation \cite{almughrabi2023pre}.
\end{enumerate}

% These images are then passed for keyframe selection based on RGB images using defocus blur \cite{de2013image}, which removes blurry images and Near image similarity \cite{idealods2019imagededup} to remove duplicate images and only keep overlapping images \cite{almughrabi2023pre}. 

% We estimate camera poses and 3D point clouds using these keyframes using PixSfM \cite{lindenberger2021pixel}. Concurrently, we employ SAM \cite{kirillov2023segment} to segment the reference object and XMem2 \cite{bekuzarov2023xmem++}, a memory-tracking-based object segmentation. Additionally, we utilize a binary image segmentation method on the RGB images, reference object masks, and food object masks to produce RGBA images. We then transform the RGBA images, poses, and PointCloud to generate meaningful metadata and create modeled data. This model data is input into NeuS2 \cite{wang2023neus2} to reconstruct colorful meshes for reference and food objects. To ensure accuracy, we remove isolated pieces from the mesh and eliminate small isolated pieces that do not belong to the food mesh. Finally, we manually identify the scaling factor using the reference mesh via MeshLab \cite{cignoni2008meshlab}, fine-tune the scaling factor using depth information and the food masks, and then apply the fine-tuned scaling factor to the cleaned food mesh.

\section{Methodology}
\label{sec:methodology}
Our study utilizes multi-view reconstruction to create intricate food meshes and calculate precise food volumes.  

\begin{figure*}[h]
    \centering
    \includegraphics[trim={0cm 0cm 0cm 2cm},clip,width=1\linewidth]{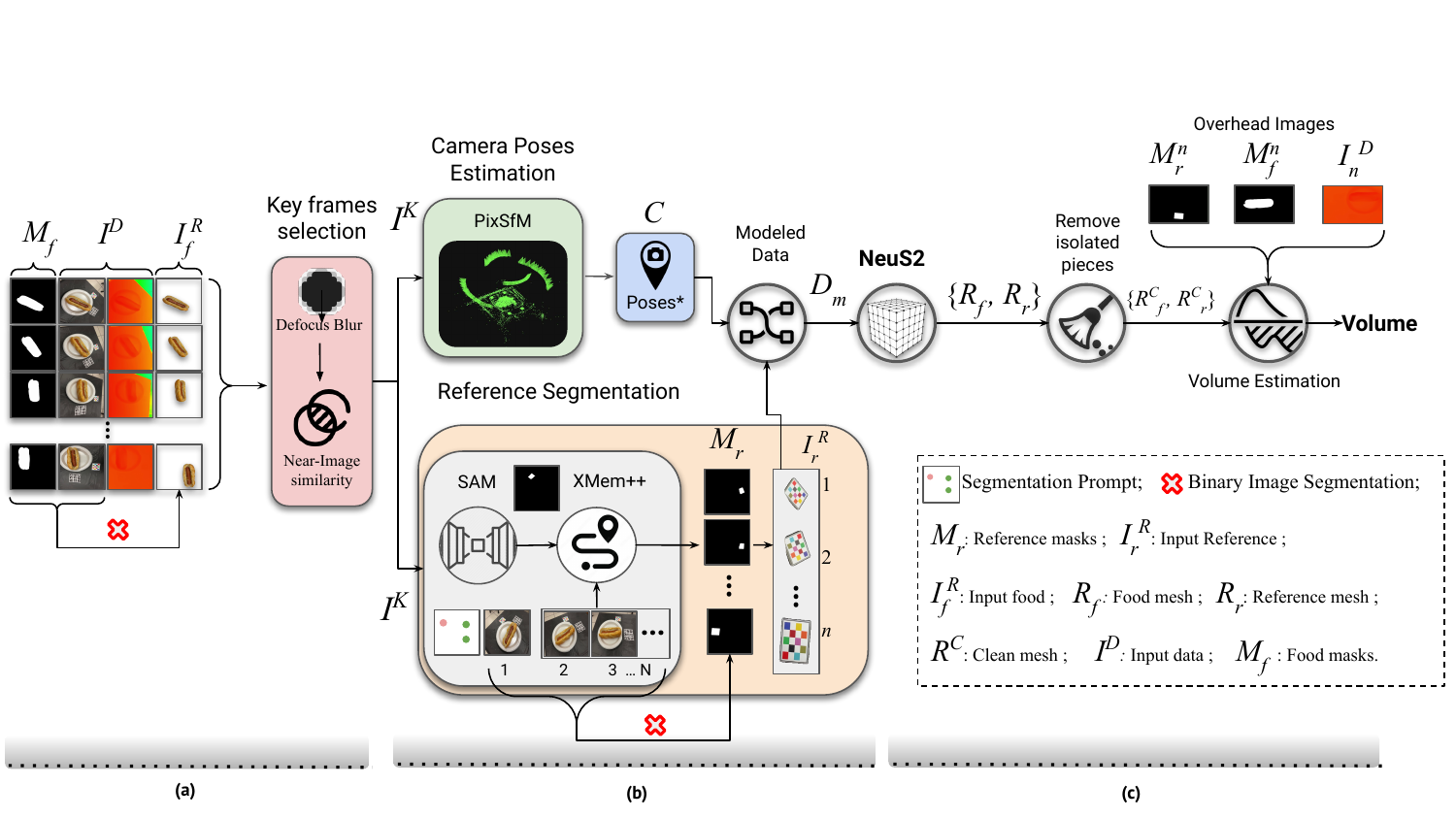}
    \caption{
Our few-shot approach for estimating food volume in (a) a few shots involves taking (\(\mathcal{I}^D\)) and food object masks as input. We start by selecting keyframes based on the RGB images, removing blurry and highly overlapped images resulting (\(I^K\)). Then, (b) we use PixSfM to estimate camera poses (\(C\)). Simultaneously, we segment the reference object using SAM with a segmentation prompt provided by a user. We then use the XMem++ method for memory-tracking to produce reference object masks for all frames, using the reference object mask and RGB images. After that, we apply a binary image segmentation method to RGB images (\(I^K\)), reference object masks (\(M_r\)), and food object masks (\(M_f\)), resulting in RGBA images (\(I^R_r\)). In contrast, we transform the RGBA images and poses to generate meaningful metadata and create modeled data (\(D_m\)). Next, (c) we input the modeled data into NeuS2 to reconstruct colorful meshes for reference (\(R_r\)) and food objects (\(R_f\)). To ensure accuracy, we use ``Remove Isolated Pieces" with diameter thresholding to clean up the mesh and remove small isolated pieces that do not belong to the reference or food mesh resulting (\(\{R^C_r, R^C_f\}\)). Finally, we manually identify the scaling factor using the reference mesh via MeshLab (\(S\)). We fine-tune the scaling factor using depth information and the food masks and then apply the fine-tuned scaling factor (\(S_f\)) to the cleaned food mesh to generate a scaled food mesh (\(R^F_f\)) in meter unit.
}
    \label{fig:methodology}
\end{figure*}

\subsection{Overview}
Our approach combines computer vision and deep learning techniques to estimate food volume from RGBD images and masks accurately. Keyframe selection ensures data quality, aided by perceptual hashing and blur detection. Camera pose estimation and object segmentation lay the groundwork for neural surface reconstruction, producing detailed meshes for volume estimation. Refinement steps enhance accuracy, including isolated piece removal and scaling factor adjustment. Our approach offers a comprehensive solution for precise food volume assessment with potential applications in nutrition analysis.

\subsection{Our Proposal: VolETA}
We begin our approach by acquiring input data, specifically RGBD images and corresponding food object masks. These RGBD images, denoted as $\mathcal{I}^D=\{{I_i^D}\}_{i=1}^n$, where \( n \) is the total number of frames, provide the necessary depth information alongside the RGB images. The food object masks, denoted as \( \{M_f^i\}_{i=1}^n \), aid in identifying the regions of interest within these images.
% Keyframe selection
Next, we proceed with keyframe selection. From the set \( \{I^D_i\}_{i=1}^n \), keyframes \( \{I_i^K\}_{j=1}^k \subseteq \{I^D_i\}_{i=1}^n \) are selected. We implement a method to detect and remove duplicates \cite{idealods2019imagededup} and blurry images \cite{de2013image} to ensure high-quality frames. This involves applying the Gaussian blurring kernel followed by the fast Fourier transform method. near-Image Similarity \cite{idealods2019imagededup} employs a perceptual hashing and hamming distance thresholding to detect similar images and keep overlapping. The duplicates and blurry images are excluded from the selection process to maintain data integrity and accuracy, as shown in Fig.~\ref{fig:methodology}(a).

% PixSfM.
Using the selected keyframes \( \{I^K_j\}_{j=1}^k \), we estimate the camera poses through PixSfM \cite{lindenberger2021pixel} (i.e., extracting features using SuperPoint \cite{detone2018superpoint}, matching them using SuperGlue \cite{sarlin2020superglue}, and refining them). The outputs are the set of camera poses \( \{C_j\}_{j=1}^k \), which are crucial for spatial understanding of the scene.
% Finding Reference mask 
In parallel, we utilize the SAM \cite{kirillov2023segment} for reference object segmentation. SAM segments the reference object with a user-provided segmentation prompt (i.e., user click), producing a reference object mask \( M^R \) for each keyframe. This mask is a foundation for tracking the reference object across all frames. We then apply the XMem++ \cite{bekuzarov2023xmem++} method for memory tracking, which extends the reference object mask \( M^R \) to all frames, resulting in a comprehensive set of reference object masks \( \{M^R_i\}_{i=1}^n \). This ensures consistency in reference object identification throughout the dataset.
% Binary Image Segmentation
To create RGBA images, we combine the RGB images, reference object masks \( \{M^R_i\}_{i=1}^n \), and food object masks \( \{M^F_i\}_{i=1}^n \). This step, denoted as \( \{I^R_i\}_{i=1}^n \), integrates the various data sources into a unified format suitable for further processing, as shown in Fig.~\ref{fig:methodology}(b).

% Transformation 
We convert the RGBA images \( \{I^R_i\}_{i=1}^n \) and camera poses \( \{C_j\}_{j=1}^k \) into meaningful metadata and modeled data \( D_m \). This transformation facilitates the accurate reconstruction of the scene. 
% NeuS2
The modeled data \( D_m \) is then input into NeuS2 \cite{wang2023neus2} for mesh reconstruction. NeuS2 generates colorful meshes \( \{R_f, R_r\} \) for the reference and food objects, providing detailed 3D representations of the scene components. We apply the ``Remove Isolated Pieces" technique to refine the reconstructed meshes. Given that the scenes contain only one food item, we set the diameter threshold to 5\% of the mesh size. This method deletes isolated connected components whose diameter is less than or equal to this 5\% threshold, resulting in a cleaned mesh \( \{R^C_f, R^C_r\} \). This step ensures that only significant and relevant parts of the mesh are retained.

%  Volume Estimation
We manually identify an initial scaling factor \( S \) using the reference mesh via MeshLab \cite{cignoni2008meshlab} for scaling factor identification. This factor is then fine-tuned \( S_{f} \) using depth information and food and reference masks, ensuring accurate scaling relative to real-world dimensions. Finally, the fine-tuned scaling factor \( S_{f} \) is applied to the cleaned food mesh \( R^C_f \), producing the final scaled food mesh \( R^F_f \). This step culminates in an accurately scaled 3D representation of the food object, enabling precise volume estimation, as shown in Fig.~\ref{fig:methodology}(c).

\subsubsection{Detecting the scaling factor}
\begin{figure}[htb]
    \centering
    \begin{subfigure}[b]{0.45\linewidth}
         \centering
            \includegraphics[width=1\linewidth, angle=90]{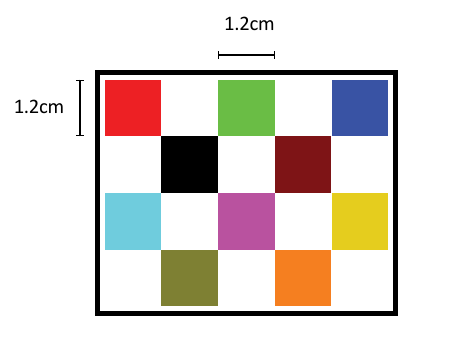}
         \caption{ }
         \label{fig:chessboard}
     \end{subfigure}
    \begin{subfigure}[b]{0.45\linewidth}
         \centering
         \includegraphics[width=1\linewidth]{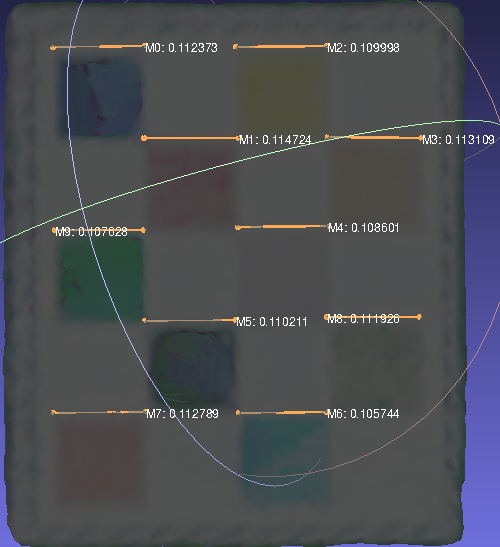}
         \caption{ }
         \label{fig:meshlab_scaling_factor}
     \end{subfigure}
    \caption{We manually measure the scaling factor using MeshLab's Measuring tool. We measure multiple blocks in the reference object mesh; then, we take the average of blocks lengths \(l_{avg}\). }
    \label{fig:scaling_factor}
\end{figure}

Generally, 3D reconstruction methods generate unitless meshes (i.e., no physical scale) by default. To overcome this limitation, we manually identify the scaling factor by measuring the distance for each block for the reference object mesh, as shown in Fig.~\ref{fig:meshlab_scaling_factor}. Next, we take the average of all blocks lengths \(l_{avg}\), while the actual real-world length (as shown in Fig.~\ref{fig:chessboard}) is constant \(l_{real}=0.012\) in meter. Furthermore, we apply the scaling factor \(S=l_{real}/l_{avg}\) on the clean food mesh \( R^C_f \), producing the final scaled food mesh \( R^F_f \) in meter.

We leverage depth information alongside food and reference object masks to validate the scaling factors. Our method for assessing food size entails utilizing overhead RGB images for each scene. Initially, we determine the pixel-per-unit (PPU) ratio (in meters) using the reference object. Subsequently, we extract the food width (\(f_{w}\)) and length (\(f_{l}\)) employing a food object mask. To ascertain the food height (\(f_{h}\)), we follow a two-step process. Firstly, we conduct binary image segmentation using the overhead depth and reference images, yielding a segmented depth image for the reference object. We then calculate the average depth utilizing the segmented reference object depth (\(d_{r}\)). Similarly, employing binary image segmentation with an overhead food object mask and depth image, we compute the average depth for the segmented food depth image (\(d_{f}\)). Finally, the estimated food height \(f_{h}\) is computed as the absolute difference between \(d_{r}\) and \(d_{f}\). Furthermore, to assess the accuracy of the scaling factor \(S\), we compute the food bounding box volume (\((f_{w} \times f_{l} \times f_{h}) \times PPU \)). We evaluate if the scaling factor \(S\) generates a food volume close to this potential volume, resulting in \(S_{fine}\). Table~\ref{tab:scaling_factors} shows the scaling factors, PPU, 2D Reference object dimensions, 3D food object dimensions, and potential volume.

\begin{table*}[h]
    \small
    \centering
    \begin{tabular}{c|c|l|c|c|cc|ccc|c}
    \hline
         Level & Id & Label & \(S_{f}\) & PPU & \multicolumn{2}{c|}{\(R_{w} \times R_{l}\)} &  \multicolumn{3}{c|}{(\(f_{w} \times f_{l} \times f_{h}\))} & Volume (\(cm^3\))   
         \\ \hline
\multirow{8}{*}{Easy} & 1 & strawberry\_2 & 0.08955223881 & 0.01786 & 320 & 360 & 238 & 257 & 2.353 & 45.91\\
& 2 & cinnamon\_bun\_1 & 0.1043478261 & 0.02347 & 236 & 274 & 363 & 419 & 2.353 & 197.07\\
& 3 & pork\_rib\_2 & 0.1043478261 & 0.02381 & 246 & 270 & 435 & 778 & 1.176 & 225.79\\
& 4 & corn\_2 & 0.08823529412 & 0.01897 & 291 & 339 & 262 & 976 & 2.353 & 216.45\\
& 5 & french\_toast\_2 & 0.1034482759 & 0.02202 & 266 & 292 & 530 & 581 & 2.53 & 377.66\\
& 6 & sandwich\_2 & 0.1276595745 & 0.02426 & 230 & 265 & 294 & 431 & 2.353 & 175.52\\
& 7 & burger\_1 & 0.1043478261 & 0.02435 & 208 & 264 & 378 & 400 & 2.353 & 211.03\\
& 8 & cake\_1 & 0.1276595745 & 0.02143 & 256 & 300 & 298 & 310 & 4.706 & 199.69\\
\hline
\multirow{7}{*}{Medium} & 9 & blueberry\_muffin & 0.08759124088 & 0.01801 & 291 & 357 & 441 & 443 & 2.353 & 149.12 \\
& 10 & banana\_2 & 0.08759124088 & 0.01705 & 315 & 377 & 446 & 857 & 1.176 & 130.80\\
& 11 & salmon\_1 & 0.1043478261 & 0.02390 & 242 & 269 & 201 & 303 & 1.176 & 40.94\\
& \cellcolor{red!25}12 & \cellcolor{red!25} steak\_1 & \cellcolor{red!25}0.127659574 & \cellcolor{red!25}0.02256 & \cellcolor{red!25}255 & \cellcolor{red!25}285 & \cellcolor{red!25}581 & \cellcolor{red!25}816 & \cellcolor{red!25}2.353 & \cellcolor{red!25}567.78\\
& 13 & burrito\_1 & 0.1034482759 & 0.02372 & 244 & 271 & 251 & 917 & 2.353 & 304.87\\
& 14 & frankfurt\_sandwich\_2 & 0.1034482759 & 0.02115 & 266 & 304 & 400 & 1022 & 2.353 & 430.29\\
& \cellcolor{red!25}15 & \cellcolor{red!25} chicken\_nugget\_2 & \cellcolor{red!25}0.08759124088 & \cellcolor{red!25}0.01328 & \cellcolor{red!25}407 & \cellcolor{red!25}484 & \cellcolor{red!25}320 & \cellcolor{red!25}337 & \cellcolor{red!25}1.256 & \cellcolor{red!25}23.89\\
\hline
\multirow{5}{*}{Hard} & 16 & everything\_bagel & 0.08759124088 & 0.01747 & 306 & 368 & 458 & 484 & 1.176 & 79.61\\
& 17 & croissant\_2 & 0.1276595745 & 0.01751 & 319 & 367 & 395 & 695 & 2.176 & 183.39 \\
& 18 & shrimp\_2 & 0.08759124088 & 0.02021 & 249 & 318 & 186 & 195 & 0.987 & 14.64\\
& 19 & waffle\_2 & 0.01034482759 & 0.01902 & 294 & 338 & 465 & 537 & 0.8 & 72.29 \\
& 20 & pizza & 0.01034482759 & 0.01913 & 292 & 336 & 442 & 651 & 1.176 & 123.97 \\
         
    \hline
    \end{tabular}
    \caption{a list of information that extracted using the RGBD and masks, where we present the scene Id, the scaling factor \(S_{fine}\), Pixel-Per-Unit (in cm), 2D reference object dimensions ($R_w \times R_l$), 3D food object dimensions ($f_w\times f_l \times f_h$) in pixels, and the potential volume (in $cm^3$). The rows in red are excluded meshes.}
    \label{tab:scaling_factors}
\end{table*}

\begin{figure}[htb]
    \centering
    \includegraphics[trim={6cm 4cm 5cm 3.5cm},clip,width=1\linewidth]{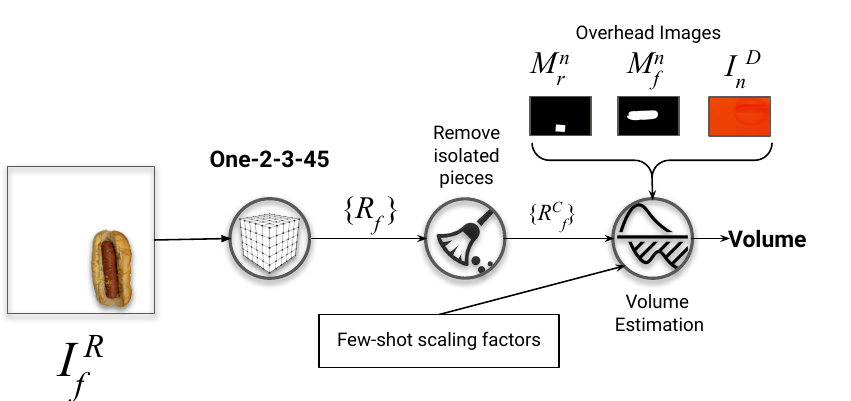}
    \caption{Our one-shot food volume estimation approach. We begin with a food-segmented image (\(I^R_f\)), and then we use the One-2-3-45 model to generate a mesh (\(R_f\)). Next, we clean up the isolated pieces that are less than 5\% of the (\(R_f\)) size, resulting in a cleaned food mesh \(R^C_f\). Furthermore, we choose a scaling factor based on the depth information \(S_f\). Finally, we apply the chosen scaling factor on \(R^C_f\) to have a scaled mesh (\(R^F_f\)) where we extract the volume.}
    \label{fig:one-shot-methodology}
\end{figure}

For one-shot 3D reconstruction, we leverage One-2-3-45 \cite{liu2024one} for reconstructing a 3D from a single RGBA view input after applying binary image segmentation on both food RGB and mask. Next, we remove isolated pieces from the generated mesh. After that, we reuse the scaling factor \(S\), which is closer to the potential volume of the clean mesh, as shown in Fig.~\ref{fig:one-shot-methodology}.

\section{Experimental Results}
\label{sec:results}
Our approach is evaluated on a 3D food dataset from the MetaFood CVPR workshop \cite{mtf_challenge}.

\begin{figure*}[h]
    \centering
    \begin{subfigure}[b]{0.16\linewidth}
         \centering
            \includegraphics[trim={5cm 2cm 5cm 3cm},clip,width=0.45\linewidth]{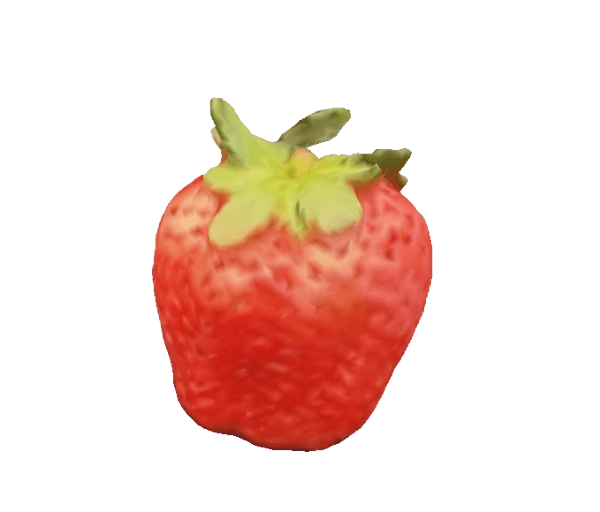}
            \includegraphics[trim={5cm 2cm 5cm 3cm},clip,width=0.45\linewidth]{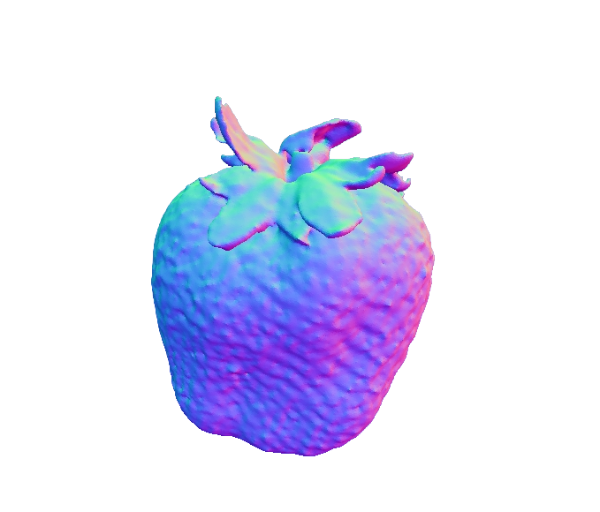}
         \caption{strawberry (1)}
         \label{fig:rsl_1}
     \end{subfigure}
     \begin{subfigure}[b]{0.16\linewidth}
         \centering
            \includegraphics[trim={5cm 5cm 5cm 5cm},clip,width=0.45\linewidth]{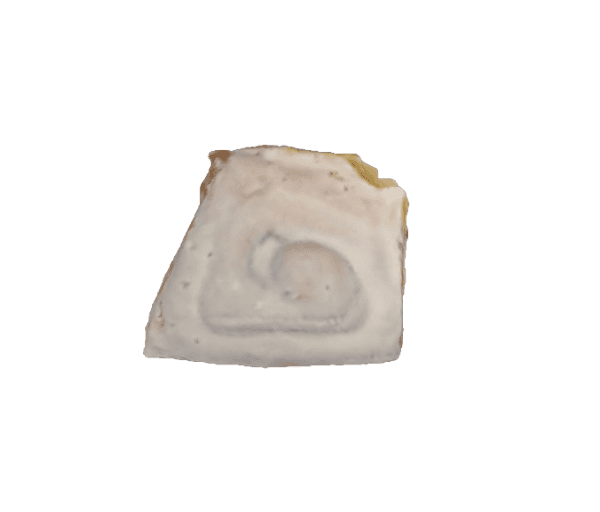}
            \includegraphics[trim={4cm 5cm 5cm 5cm},clip,width=0.45\linewidth]{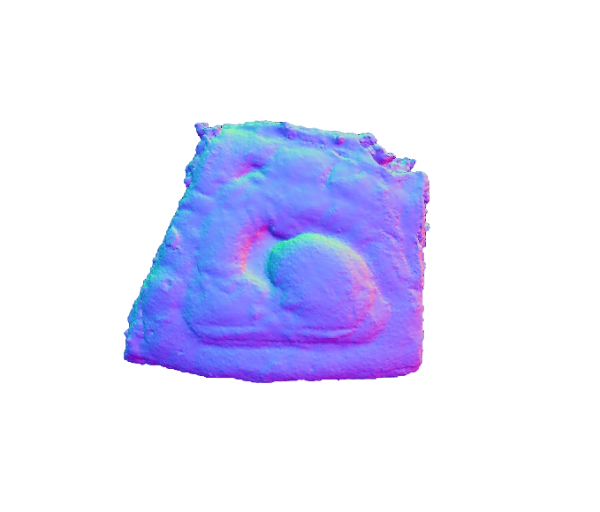}
         \caption{cinnamon bun (2)}
         \label{fig:rsl_2}
     \end{subfigure}
     \begin{subfigure}[b]{0.16\linewidth}
         \centering
            \includegraphics[trim={5cm 3cm 5cm 3cm},clip,width=0.45\linewidth]{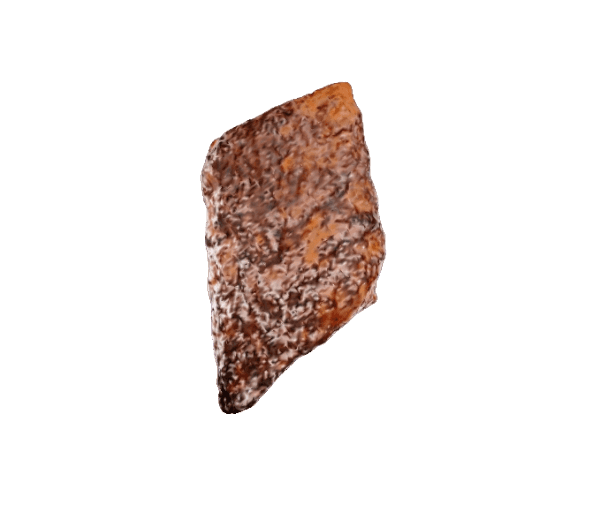}
            \includegraphics[trim={5cm 1cm 5cm 3cm},clip,width=0.45\linewidth]{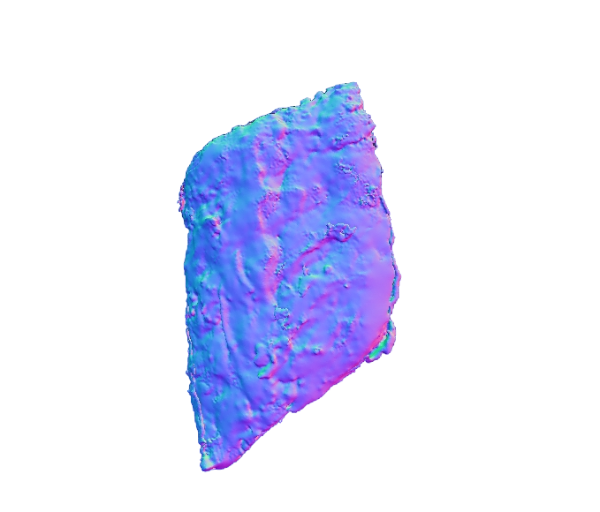}
         \caption{pork rib (3)}
         \label{fig:rsl_3}
     \end{subfigure}
      \begin{subfigure}[b]{0.16\linewidth}
         \centering
            \includegraphics[trim={5cm 3cm 5cm 3cm},clip,width=0.45\linewidth]{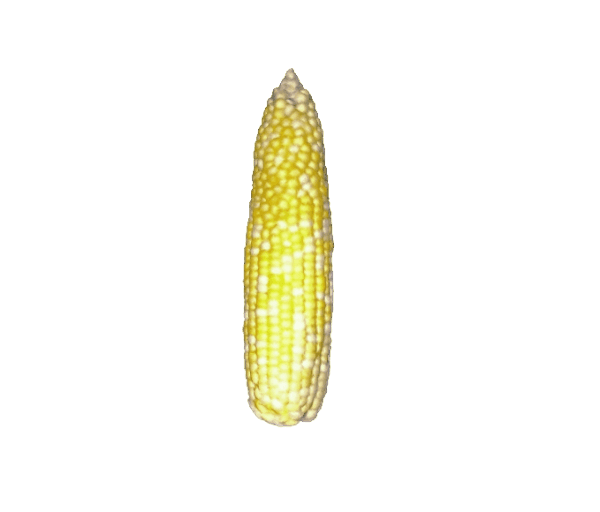}
            \includegraphics[trim={5cm 3cm 5cm 3cm},clip,width=0.45\linewidth]{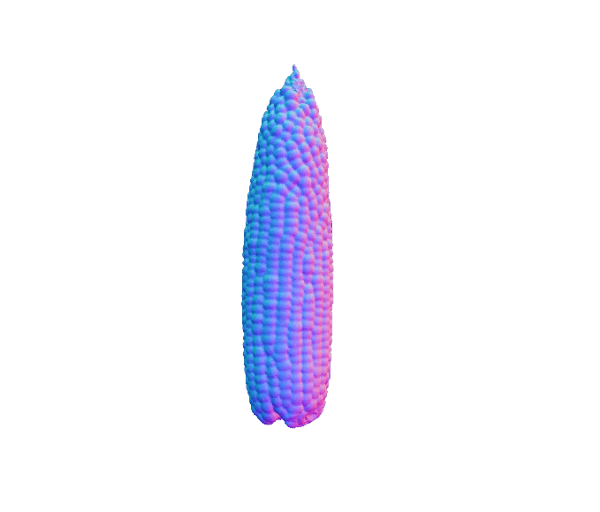}
         \caption{corn (4)}
         \label{fig:rsl_4}
     \end{subfigure}
     \begin{subfigure}[b]{0.16\linewidth}
         \centering
            \includegraphics[trim={5cm 5cm 5cm 5cm},clip,width=0.45\linewidth]{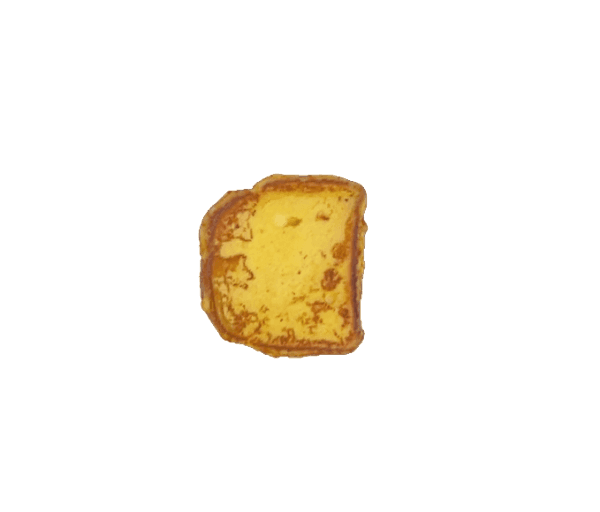}
            \includegraphics[trim={5cm 4.5cm 5cm 5cm},clip,width=0.45\linewidth]{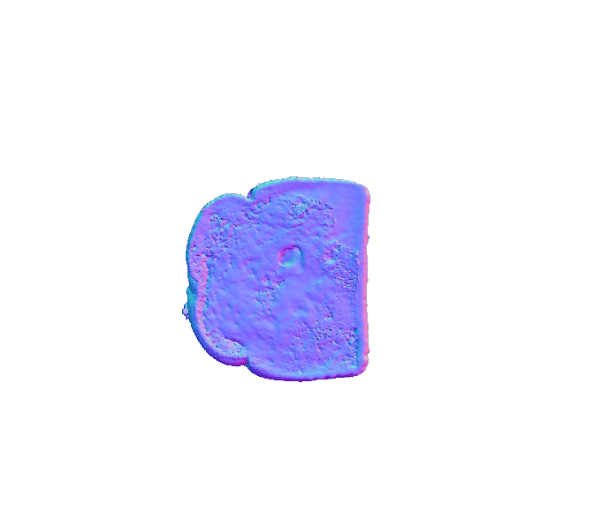}
         \caption{french toast (5)}
         \label{fig:rsl_5}
     \end{subfigure}
     \begin{subfigure}[b]{0.16\linewidth}
         \centering
            \includegraphics[trim={5cm 4cm 5cm 5cm},clip,width=0.45\linewidth]{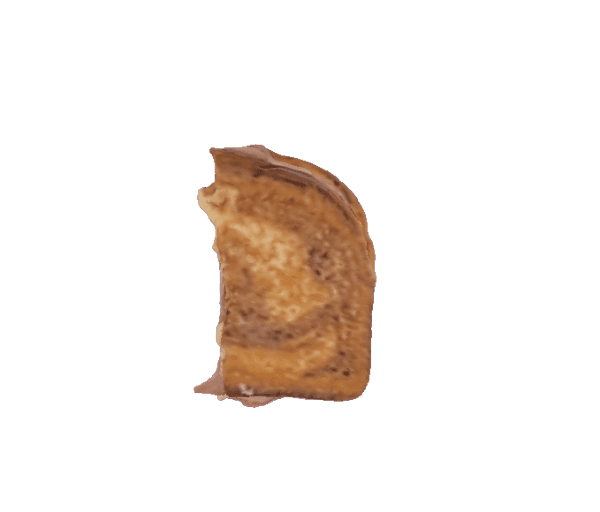}
            \includegraphics[trim={5cm 4cm 5cm 5cm},clip,width=0.45\linewidth]{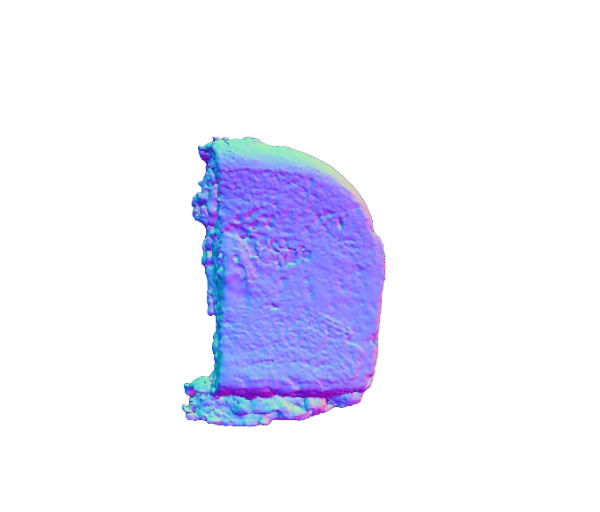}
         \caption{sandwich (6)}
         \label{fig:rsl_6}
     \end{subfigure}
     \begin{subfigure}[b]{0.16\linewidth}
         \centering
            \includegraphics[trim={6cm 5cm 6cm 6cm},clip,width=0.45\linewidth]{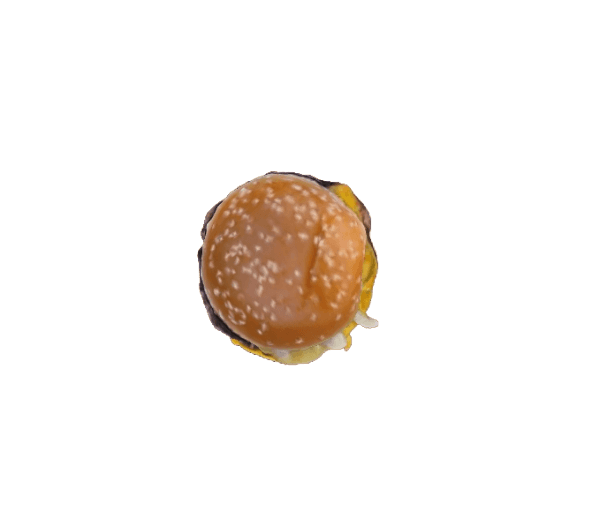}
            \includegraphics[trim={6cm 5cm 6cm 6cm},clip,width=0.45\linewidth]{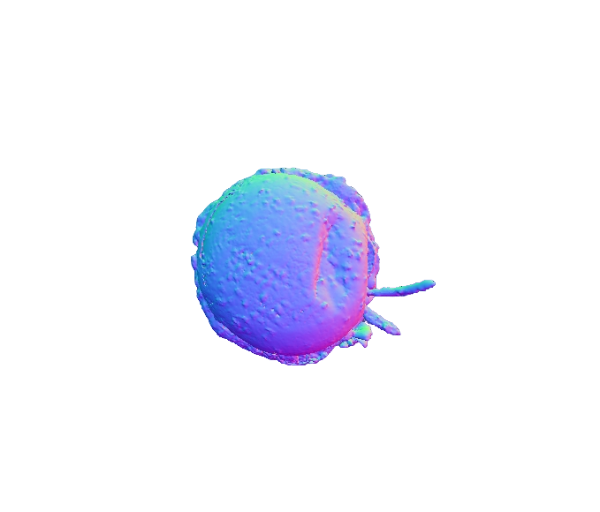}
         \caption{burger (7)}
         \label{fig:rsl_7}
     \end{subfigure}
      \begin{subfigure}[b]{0.16\linewidth}
         \centering
            \includegraphics[trim={6cm 6cm 6cm 6cm},clip,width=0.45\linewidth]{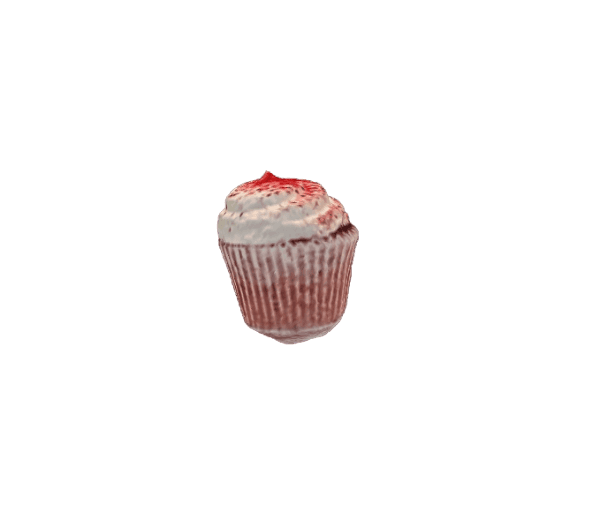}
            \includegraphics[trim={6cm 6cm 6cm 6cm},clip,width=0.45\linewidth]{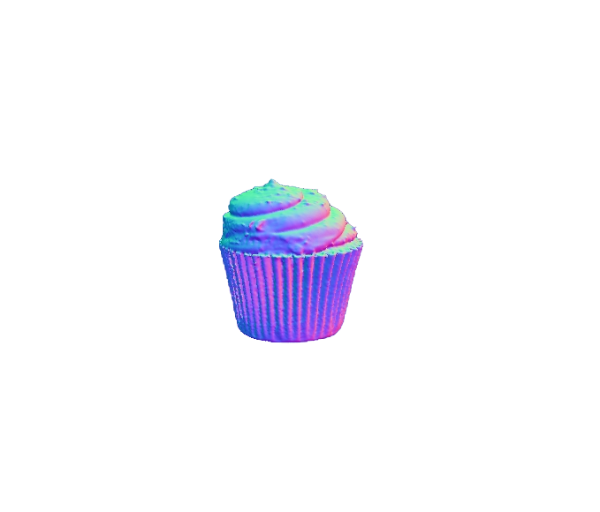}
         \caption{cake (8)}
         \label{fig:rsl_8}
     \end{subfigure}
     \begin{subfigure}[b]{0.16\linewidth}
         \centering
            \includegraphics[trim={5cm 5cm 5cm 5cm},clip,width=0.45\linewidth]{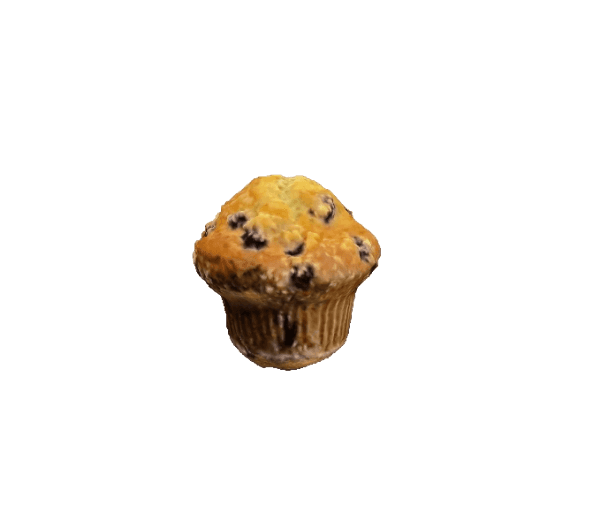}
            \includegraphics[trim={5cm 5cm 5cm 5cm},clip,width=0.45\linewidth]{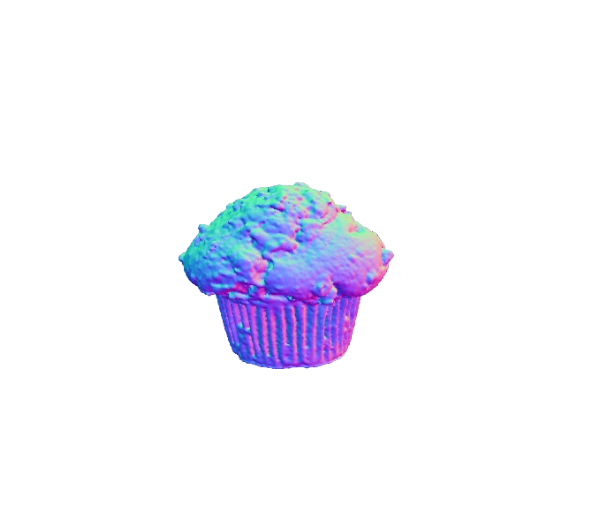}
         \caption{blueberry muffin (9)}
         \label{fig:rsl_9}
     \end{subfigure}
     \begin{subfigure}[b]{0.16\linewidth}
         \centering
            \includegraphics[trim={5cm 2cm 5cm 3cm},clip,width=0.45\linewidth]{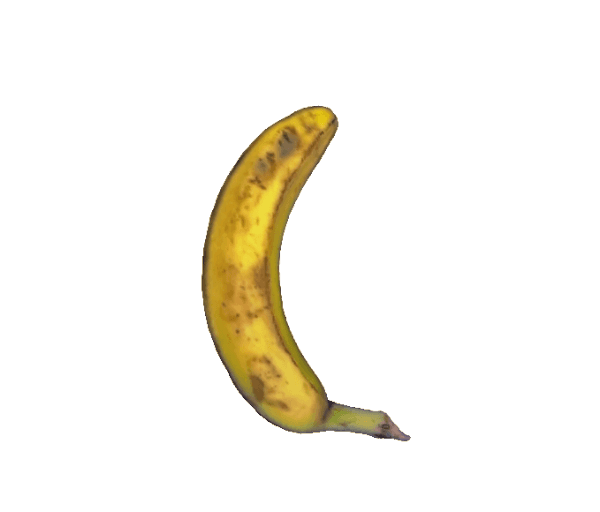}
            \includegraphics[trim={5cm 2cm 5cm 3cm},clip,width=0.45\linewidth]{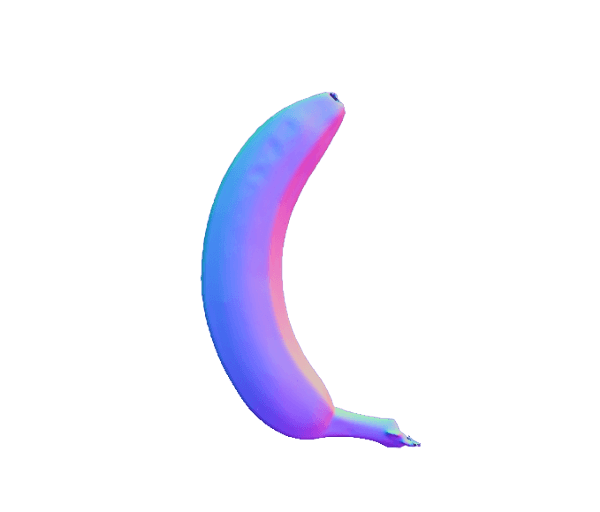}
         \caption{banana (10)}
         \label{fig:rsl_10}
     \end{subfigure}
     \begin{subfigure}[b]{0.16\linewidth}
         \centering
            \includegraphics[trim={6cm 5cm 6cm 6cm},clip,width=0.45\linewidth]{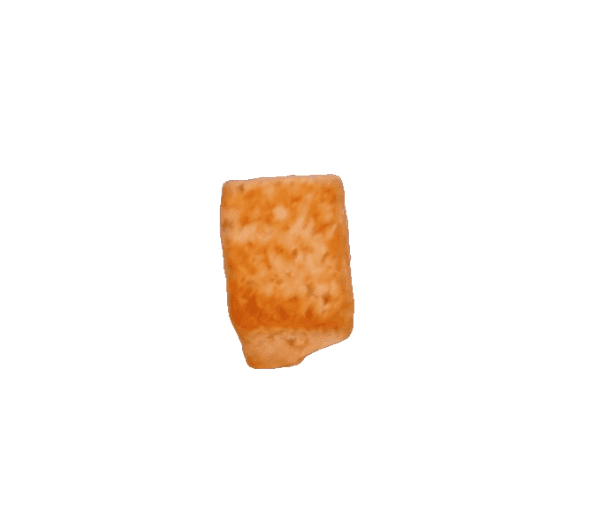}
            \includegraphics[trim={6cm 4.5cm 6cm 6cm},clip,width=0.45\linewidth]{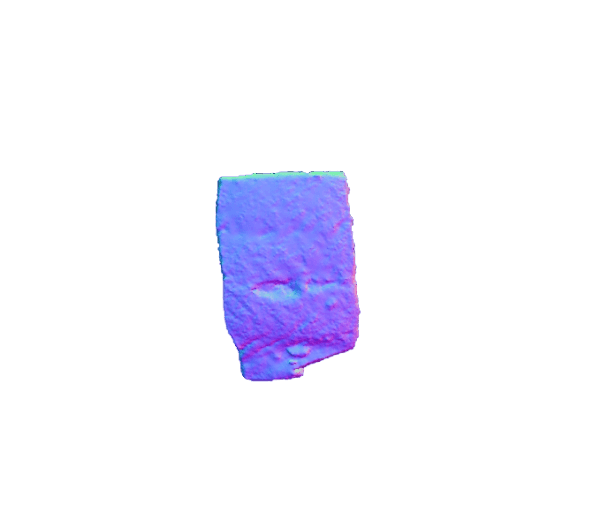}
         \caption{salmon (11)}
         \label{fig:rsl_11}
     \end{subfigure}
     \begin{subfigure}[b]{0.16\linewidth}
         \centering
            \includegraphics[trim={5cm 5cm 5cm 5cm},clip,angle=90,width=0.45\linewidth]{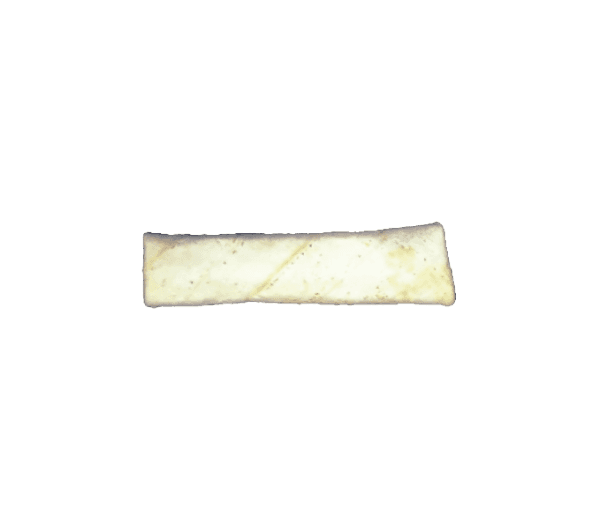}
            \includegraphics[trim={5cm 5cm 5cm 5cm},clip,angle=90,width=0.45\linewidth]{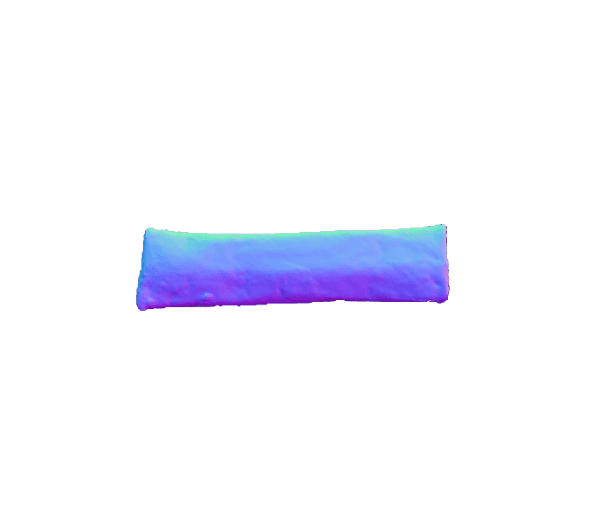}
         \caption{burrito (13)}
         \label{fig:rsl_13}
     \end{subfigure}
     \begin{subfigure}[b]{0.16\linewidth}
         \centering
            \includegraphics[trim={5cm 2cm 5cm 3cm},clip,width=0.45\linewidth]{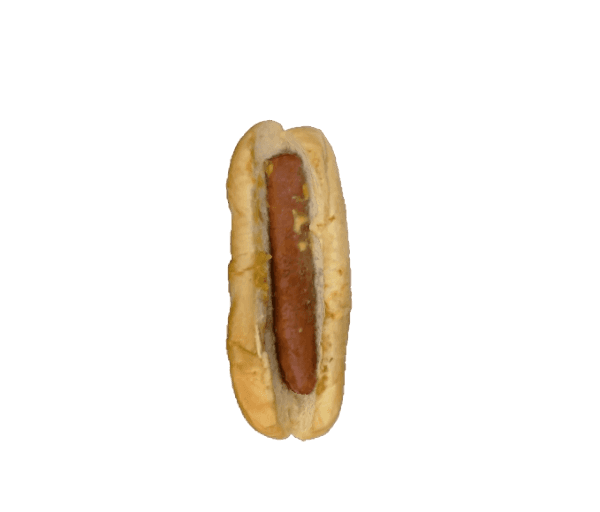}
            \includegraphics[trim={5cm 2cm 5cm 3cm},clip,width=0.45\linewidth]{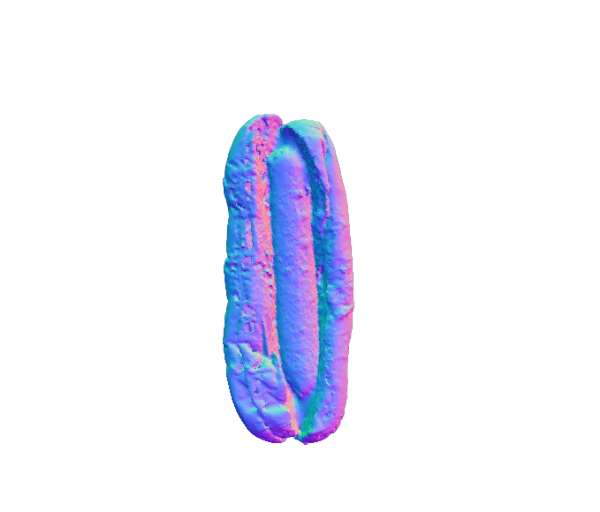}
         \caption{frank sandwich (14)}
         \label{fig:rsl_14}
     \end{subfigure}
     \begin{subfigure}[b]{0.16\linewidth}
         \centering
            \includegraphics[trim={7cm 6cm 6cm 6cm},clip,width=0.45\linewidth]{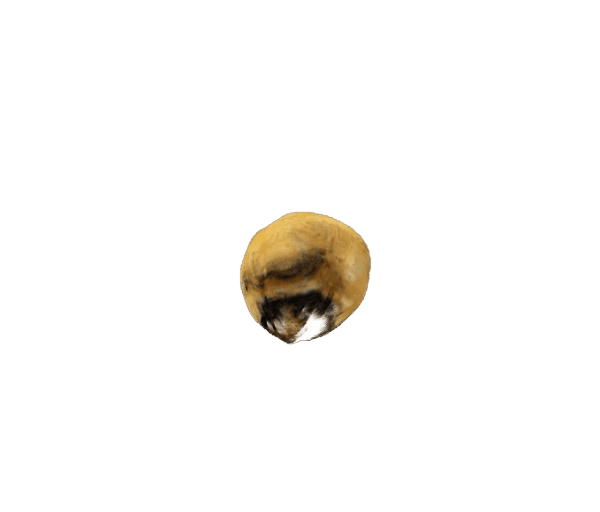}
            \includegraphics[trim={6cm 5cm 6cm 6cm},clip,width=0.45\linewidth]{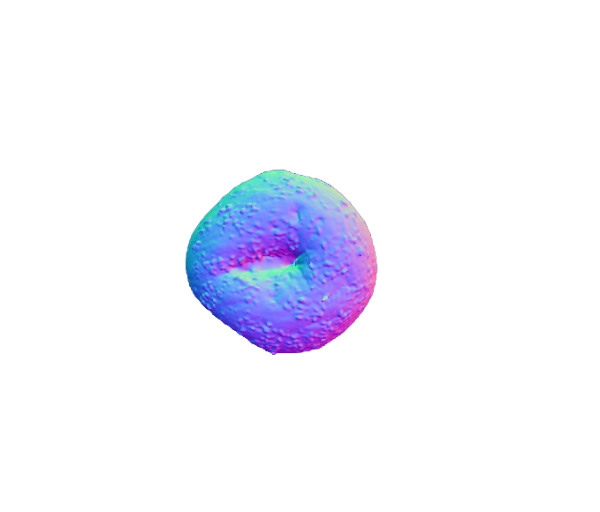}
         \caption{everything bagel (16)}
         \label{fig:rsl_16}
     \end{subfigure}
     \begin{subfigure}[b]{0.16\linewidth}
         \centering
            \includegraphics[trim={5cm 4cm 5cm 5cm},clip,width=0.45\linewidth]{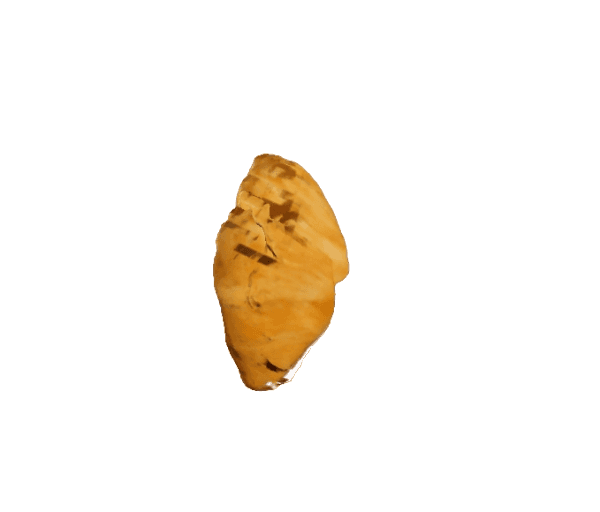}
            \includegraphics[trim={5cm 4cm 5cm 5cm},clip,width=0.45\linewidth]{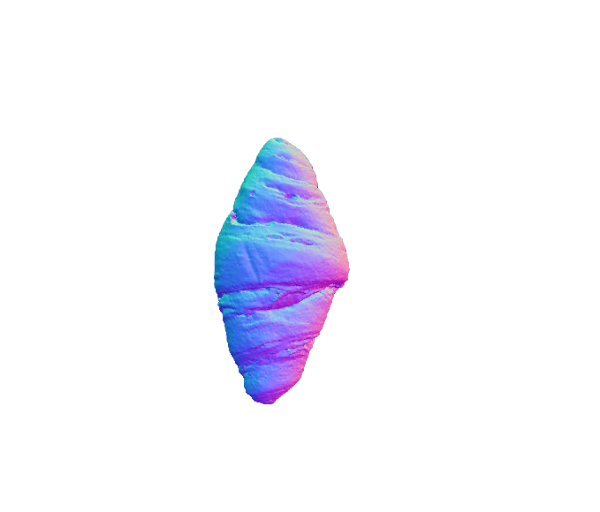}
         \caption{croissant (17)}
         \label{fig:rsl_17}
     \end{subfigure}
     \begin{subfigure}[b]{0.16\linewidth}
         \centering
            \includegraphics[trim={5cm 5cm 5cm 5cm},clip,width=0.45\linewidth]{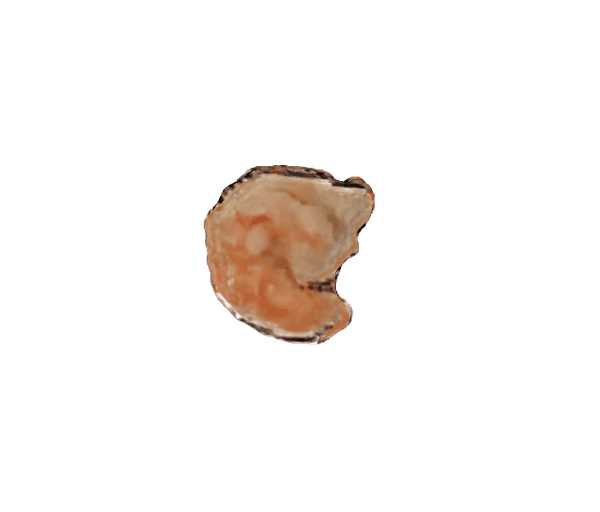}
            \includegraphics[trim={5cm 5cm 5cm 5cm},clip,width=0.45\linewidth]{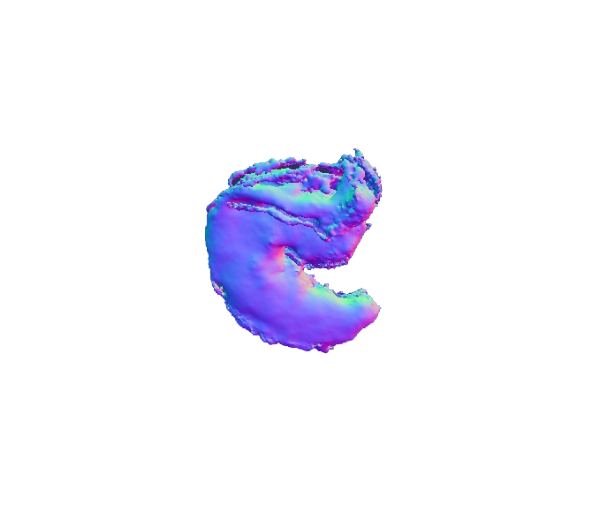}
         \caption{shrimp (18)}
         \label{fig:rsl_18}
     \end{subfigure}
     \begin{subfigure}[b]{0.16\linewidth}
         \centering
            \includegraphics[trim={5cm 5cm 5cm 5cm},clip,width=0.45\linewidth]{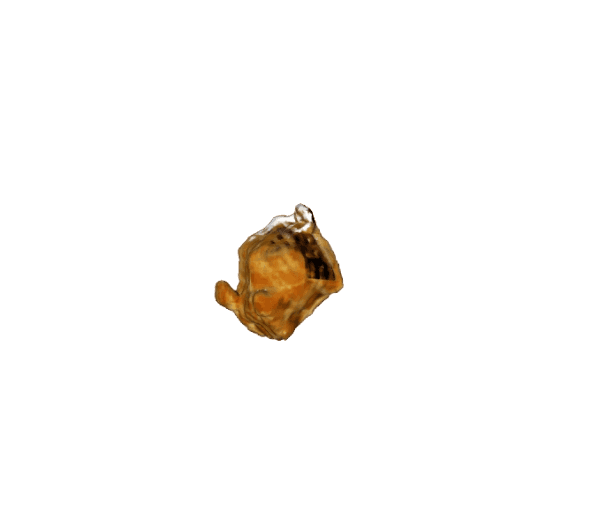}
            \includegraphics[trim={5cm 5cm 5cm 5cm},clip,width=0.45\linewidth]{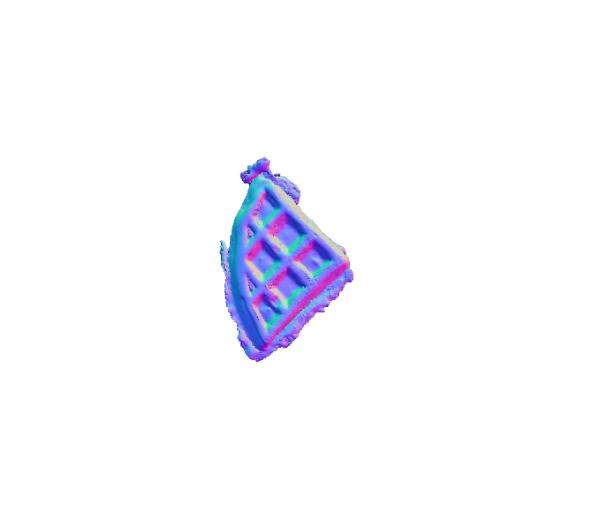}
         \caption{waffle (19)}
         \label{fig:rsl_19}
     \end{subfigure}
     \begin{subfigure}[b]{0.16\linewidth}
         \centering
            \includegraphics[trim={5cm 4cm 5cm 5cm},clip,width=0.45\linewidth]{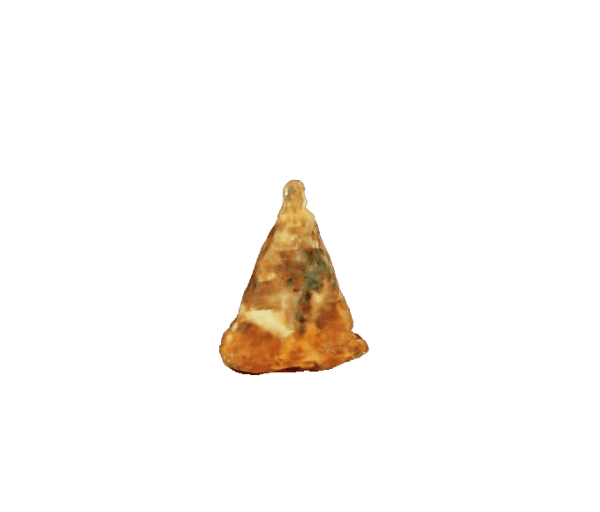}
            \includegraphics[trim={5cm 4cm 5cm 4.5cm},clip,width=0.45\linewidth]{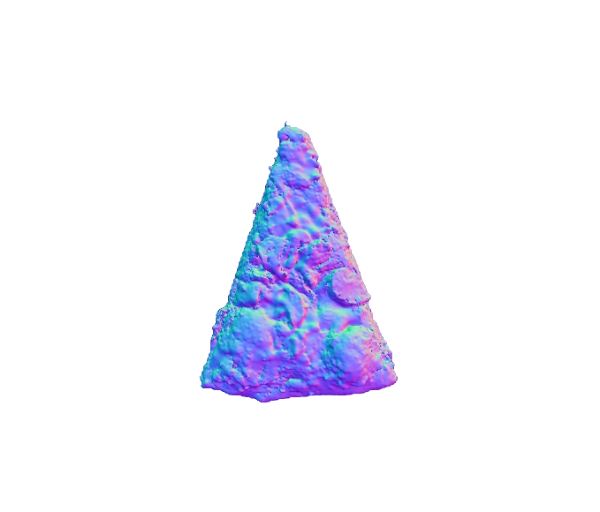}
         \caption{pizza (20)}
         \label{fig:rsl_20}
     \end{subfigure}
    \caption{
    Comparisons to ours and ground truth using the MTF dataset. Each scene shows our reconstruction (left) and ground truth (right).   
    }
    \label{fig:qualitative_results}
\end{figure*}

\subsection{Evaluation protocol}
The evaluation process consists of two distinct phases, each focusing on different aspects of the reconstructed 3D models: portion size (volume) and shape (3D structure). In Phase I, the Mean Absolute Percentage Error (MAPE) metric assesses the model's portion size accuracy. The MAPE measures the accuracy of volume predictions by comparing the predicted volume to the true volume. The formula for MAPE is:
\[ \text{MAPE} = \frac{1}{n} \sum_{i=1}^{n} \left| \frac{V_{\text{true}, i} - V_{\text{pred}, i}}{V_{\text{true}, i}} \right| \times 100\% \]
where \( V_{\text{true}, i} \) and \( V_{\text{pred}, i} \) represent the true and predicted volumes for the \( i \)-th model, respectively. This phase focuses solely on evaluating the volume accuracy of the models to determine their precision in estimating portion sizes.
In Phase II, the complete 3D mesh files for each food item validate the volume accuracy results obtained in Phase I. Subsequently, the shape accuracy of these models is evaluated using the Chamfer distance \cite{barrow1977parametric} metric. The Chamfer distance measures the similarity between two sets of points and is used to compare the 3D structure of the reconstructed models against the ground truth. This two-phase evaluation process ensures a comprehensive assessment of both the portion size and the structural accuracy of the 3D food models.

\subsection{Implementation settings}
\label{sec:resource_limitation}
We ran the experiments using two GPUs, GeForce GTX 1080 Ti/12G and RTX 3060/6G. We set the hamming distance as 12 for the near image similarity. For Gaussian kernel radius, we set the even numbers in the range $[0 ... 30]$ for detecting blurry images. We set the diameter as 5\% for removing isolated pieces. The number of iteration of NeuS2 is 15000, mesh resolution is $512 \times 512$, the unit cube ``aabb\_scale" is $1$, ``scale": 0.15, and ``offset": $[0.5, 0.5, 0.5]$ for each food scene.

\subsection{MTF Dataset}
The dataset used in this study consists of 20 food scenes categorized into three levels of difficulty: easy, medium, and hard. The easy category includes the first eight scenes, each containing approximately 200 images. The medium category comprises seven scenes, each comprising around 30 images. The hard category consists of the remaining scenes, each containing only a single image. Each image in the dataset is accompanied by corresponding food masks and depth images. Additionally, each scene features a single food object, a reference board (e.g., a chessboard), and QR code papers surrounding the object. A metadata file is also provided for all scenes, describing the food object in detail. With its varying difficulty levels, this dataset structure enables a thorough evaluation of the models' performance across different scenarios, ensuring their robustness and generalizability. The dataset is available at \footnote{https://www.kaggle.com/competitions/cvpr-metafood-3d-food-reconstruction-challenge/data}.

\subsection{VolETA Results}
\begin{figure}[htb]
    \centering
    \includegraphics[width=1\linewidth]{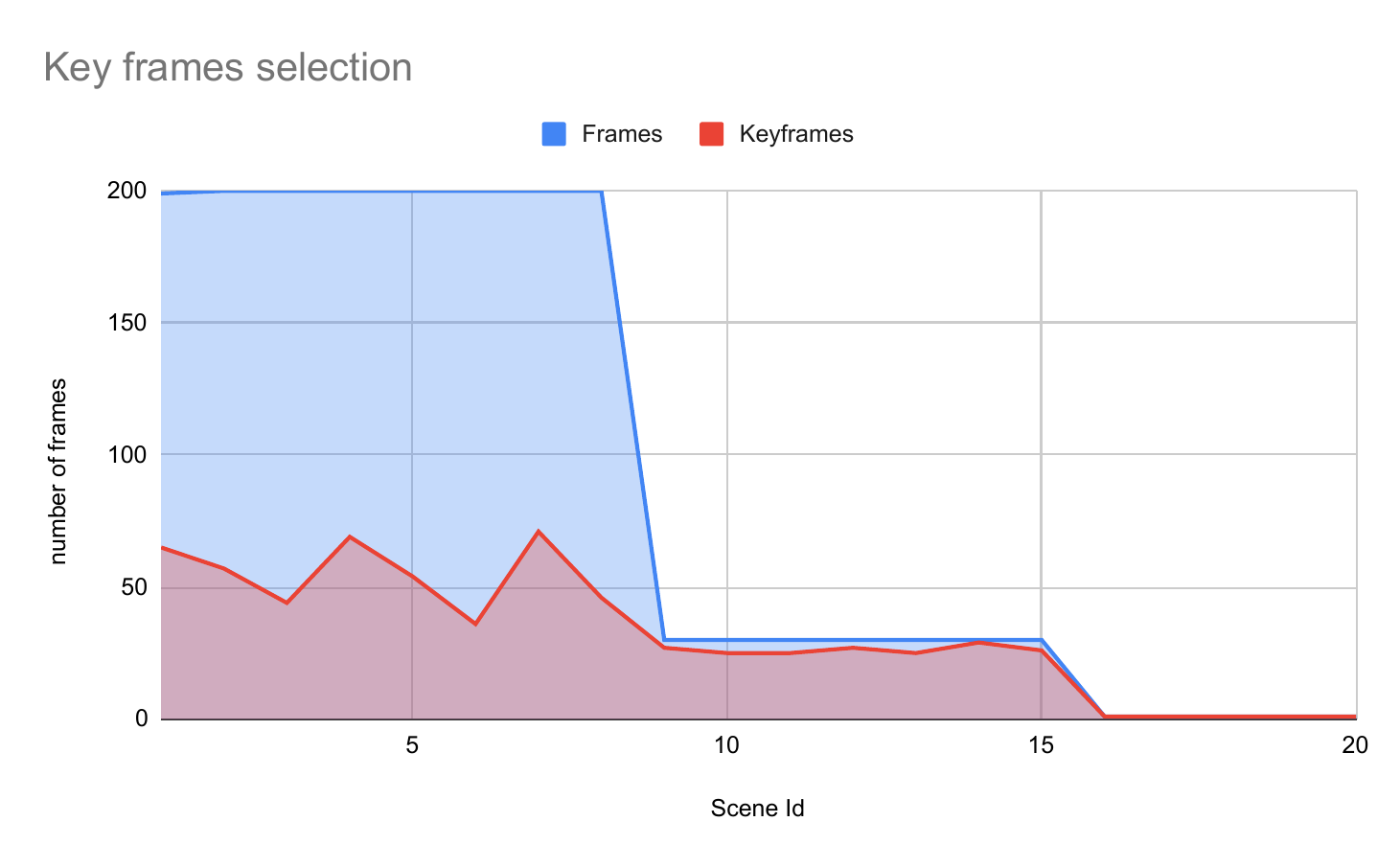}
    \caption{A quantitative results to the number of frames before and after the keyframe selection phase. Our approach is only using 34.8\% of the data.}
    \label{fig:keyframe_selection}
\end{figure}

\begin{figure}[htb]
    \centering
     \begin{subfigure}[b]{0.49\linewidth}
         \centering
            \includegraphics[width=1\linewidth]{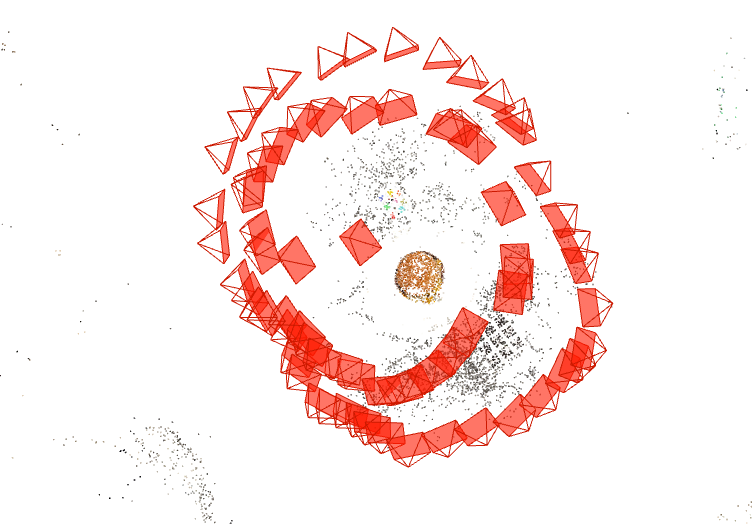}
         \caption{burger (7)}
         \label{fig:burger_1}
     \end{subfigure}
     \begin{subfigure}[b]{0.49\linewidth}
         \centering
            \includegraphics[width=1\linewidth]{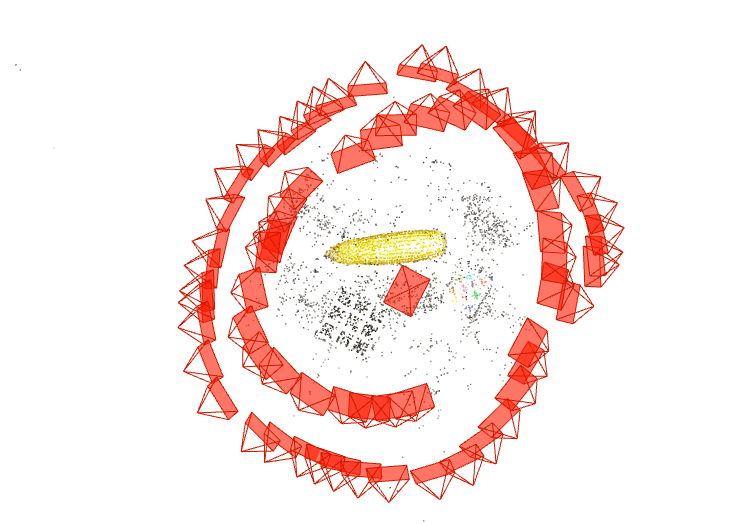}
         \caption{corn (4)}
         \label{fig:corn_2}
     \end{subfigure}
    \caption{PixSfm results after applying keyframes selection. PixSfM excels in estimating and refining camera poses by providing a rich point cloud using Superpoint feature extractors.}
    \label{fig:pixsfm}
\end{figure}

\begin{figure}[htb]
    \centering
        \includegraphics[width=1\linewidth]{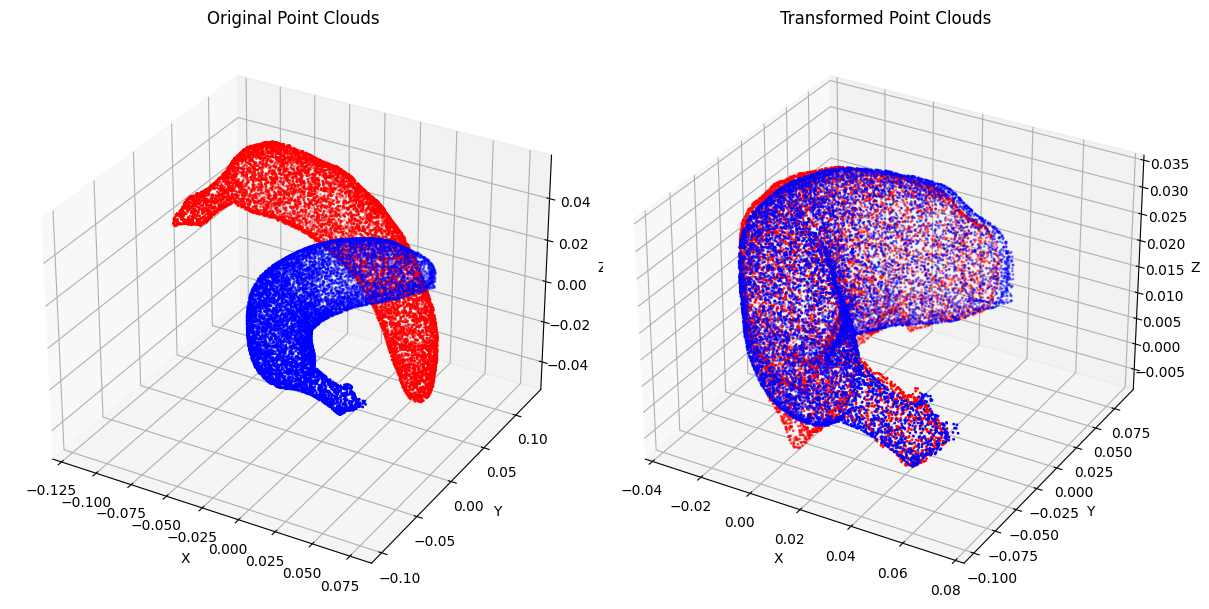}
    \caption{Performed ICP mesh registration between our generated and ground truth meshes for banana\_2 scene. Unregistered meshes are on the left, while registered meshes are on the right. Our Point cloud is red, and the ground truth is blue.}
    \label{fig:icp}
\end{figure}

We extensively validated our approach on an MTF dataset and compared our results with ground truth meshes using MAPE and Chamfer distance metrics. More Briefly, we leverage our approach for each food scene separately. A one-shot food volume estimation approach is applied if the number of keyframes $k$ equals 1. Otherwise, a few-shot food volume estimation is applied. Notably, Fig.~\ref{fig:keyframe_selection} shows that our keyframe selection process chooses 34.8\% of total frames for the rest of the pipeline, where it shows the minimum frames with the highest information.
After finding the keyframes, PixSfM \cite{lindenberger2021pixel} estimates the poses and point cloud (see Fig.~\ref{fig:pixsfm}).
After generating the scaled meshes, we calculate the volumes and Chamfer distance with and without transformation metrics. We registered our meshes and ground truth meshes to obtain the transformation metrics using ICP \cite{rusinkiewicz2001efficient} (see Fig.~\ref{fig:icp}).
Table ~\ref{tab:results_1} presents the quantitative comparisons of our volumes and Chamfer distance with and without the estimated transformation metrics from ICP.  
\begin{table}[htb]
    \small
    \centering
    \begin{tabular}{c|c|c|c|c|c}
    \hline
    L & Id & Our Vol. & GT Vol. & Ch. w/ t.m & Ch. w/o t.m  \\
    \hline
\multirow{8}{*}{E}   & 1 & 40.06 & 38.53 & 1.63 & 85.40\\
                        & 2 & 216.9 & 280.36 & 7.12 & 111.47\\
                        & 3 & 278.86 & 249.67 & 13.69 & 172.88\\
                        & 4 & 279.02 & 295.13 & 2.03 & 61.30\\
                        & 5 & 395.76 & 392.58 & 13.67 & 102.14\\
                        & 6 & 205.17 & 218.44 & 6.68 & 150.78\\
                        & 7 & 372.93 & 368.77 & 4.70 & 66.91\\
                        & 8 & 186.62 & 173.13 & 2.98 & 152.34\\
\hline
                        
\multirow{7}{*}{M} &9 & 224.08 & 232.74 & 3.91 & 160.07\\
                        & 10 & 153.76 & 163.09 & 2.67 & 138.45\\
                        & 11 & 80.4 & 85.18 & 3.37 & 151.14\\
                        & \cellcolor{red!25}12 & \cellcolor{red!25}- & \cellcolor{red!25}- & \cellcolor{red!25}- & \cellcolor{red!25}- \\
                        & 13 & 363.99 & 308.28 & 5.18 & 147.53\\
                        & 14 & 535.44 & 589.83 & 4.31 & 89.66\\
                        & \cellcolor{red!25}15 & \cellcolor{red!25}- & \cellcolor{red!25}- & \cellcolor{red!25}- & \cellcolor{red!25}- \\
\hline
\multirow{7}{*}{H}& 16& 163.13 & 262.15 & 18.06 & 28.33\\
                        & 17 & 224.08 & 181.36 & 9.44 & 28.94\\
                        & 18 & 25.4 & 20.58 & 4.28 & 12.84\\
                        & 19 & 110.05 & 108.35 & 11.34 & 23.98\\
                        & 20 & 130.96 & 119.83 & 15.59 & 31.05\\
    \hline
    \end{tabular}
    \caption{Quantitative comparison of our approach with ground truth using MTF dataset. We evaluate our approach using Chamfer distance in $\times 10^{-3}$ with and without transformation metrics. Volumes in the table is in $cm^3$.}
    \label{tab:results_1}
\end{table}
For overall our method performance, Table~\ref{tab:results_2} shows the MAPE and Chamfer distance with and without transformation metrics. 
\begin{table}[htb]
    \small
    \centering
    \begin{tabular}{c|cc|cc}
    \hline
    MAPE ↓ (\%) & \multicolumn{2}{c|}{Ch. w/ t.m ↓} & \multicolumn{2}{c}{Ch. w/o t.m ↓} \\
    & sum & mean & sum & mean \\
    \hline
    %\textbf{10.973} & 0.13066 & \textbf{0.00726} & 1.71521 & \textbf{0.09529} \\
    \textbf{10.973} & 0.130 & \textbf{0.007} & 1.715 & \textbf{0.095} \\
    \hline
    \end{tabular}
    \caption{Quantitative comparison of our approach with ground truth using MTF dataset. We evaluate our approach using Chamfer distance with and without transformation metrics. The results show the mean and sum of the 18 scenes.}
    \label{tab:results_2}
\end{table}
% waiting for Umair to finish the results.
Additionally, Fig.~\ref{fig:qualitative_results} shows the qualitative results on the one- and few-shot 3D reconstruction from the MTF dataset. The figures show that our model excels in texture details, artifact correction, missing data handling, and color adjustment across different scene parts.  

\paragraph{Limitations.}
Despite the promising results demonstrated by our method, several limitations need to be addressed in future work:
\begin{itemize}
    \item \textbf{Manual processes:} The current pipeline includes manual steps, such as providing a segmentation prompt and identifying scaling factors. These steps should be automated to enhance efficiency and reduce human intervention. This limitation arises from the necessity of using a reference object to compensate for missing data sources, such as Inertial Measurement Unit (IMU) data.
    \item \textbf{Input requirements:} Our method requires extensive input information, including food masks and depth data. Streamlining the necessary inputs would simplify the process and potentially increase its applicability in varied settings.
    \item \textbf{Complex backgrounds and objects:} We have not tested our method in environments with complex backgrounds or on highly intricate food objects. Applying our approach to datasets with more complex food items, such as the Nutrition5k \cite{thames2021nutrition5k} dataset, would be challenging and could help identify corner cases that need to be addressed.
    \item \textbf{Capturing complexities:} The method has not been evaluated under different capturing complexities, such as varying distances between the camera and the food object, different camera speeds, and other scenarios as defined in the Fruits and Vegetables \cite{steinbrener2023learning} dataset. These factors could significantly impact the performance and robustness of our method.
    \item \textbf{Pipeline complexity:} For one-shot neural rendering, we currently use the One-2-3-45 \cite{liu2024one} method. However, we aim to integrate a similar 2D diffusion model mechanism, Zero123 \cite{liu2023zero}, into our pipeline to reduce complexity and improve the efficiency of our approach.
\end{itemize}

\section{Acknowldgemet}
This work was partially funded by the EU project MUSAE (No. 01070421), 2021-SGR-01094 (AGAUR), Icrea Academia’2022 (Generalitat de Catalunya), Robo STEAM (2022-1-BG01-KA220-VET000089434, Erasmus+ EU), DeepSense (ACE053/22/000029, ACCIÓ), CERCA Programme/Generalitat de Catalunya, and Grants PID2022141566NB-I00 (IDEATE), PDC2022-133642-I00 (DeepFoodVol), and CNS2022-135480 (A-BMC) funded by MICIU/AEI/10.13039/501100 011033, by FEDER (UE), and by European Union NextGenerationEU/ PRTR. R. Marques acknowledges the support of the Serra Húnter Programme. A. AlMughrabi acknowledges the support of FPI Becas, MICINN, Spain.
\section{Conclusion}
\label{sec:conclusion}
We presented VolETA, a practical approach to measuring food volume accurately using one- and few-shot RGBD images. Our approach involves selecting keyframes using near-image similarity techniques and eliminating blurry images. We used PixSfM for camera pose estimation, point cloud generation, and the SAM model for reference object segmentation. We achieved consistent tracking of the reference object using the XMem2 method. We seamlessly integrated the data by converting the combined RGB, reference object, and food object masks into RGBA images, which led to accurate neural surface reconstruction with NeuS2. We improved the meshes using the ``Remove Isolated Pieces" technique and made precise scaling adjustments, resulting in realistic 3D representations of food objects. In addition to our few-shot 3D reconstruction approach, we explored one-shot 3D reconstruction, which involved similar keyframe selection followed by binary image segmentation, the One-2-3-45 model, ``Remove Isolated Pieces," and volume estimation. Our methods demonstrate high accuracy and efficiency, making them suitable for practical nutrition analysis and dietary monitoring applications. This integrated approach combines advanced computer vision and deep learning techniques, providing a powerful tool for precise volumetric analysis.
{
    \small
    \bibliographystyle{ieeenat_fullname}
    \bibliography{main}
}

% WARNING: do not forget to delete the supplementary pages from your submission 
% \input{sec/X_suppl}

\end{document}